\documentclass{article}
\usepackage{spconf,amsmath,amssymb,graphicx}
\usepackage{xspace}
\usepackage{booktabs}
\usepackage{graphicx}
\usepackage{multirow}

\newcommand{\ie}{\emph{i.e.}\xspace}
\newcommand{\eg}{\emph{e.g.}\xspace}
\newcommand{\etal}{\emph{et al.}\xspace}
\newcommand{\etc}{\emph{etc.}\xspace}

\title{Improving End-to-End Speech Recognition with Policy Learning}
\name{Yingbo Zhou, Caiming Xiong, Richard Socher}
\address{Salesforce Research}
\begin{document}
%\ninept
%
\maketitle
\begin{abstract}
Connectionist temporal classification (CTC) is widely used for maximum likelihood learning in end-to-end speech recognition models. However, there is usually a disparity between the negative maximum likelihood and the performance metric used in speech recognition, \eg, word error rate (WER). This results in a mismatch between the objective function and metric during training. % JB: these two sentences are mostly redundant with each other
We show that the above problem can be mitigated by jointly training with maximum likelihood and policy gradient. In particular, with policy learning we are able to directly optimize on the (otherwise non-differentiable) performance metric. We show that joint training improves relative performance by 4\% to 13\% for our end-to-end model as compared to the same model learned through maximum likelihood. The model achieves 5.53\% WER on Wall Street Journal dataset, and 5.42\% and 14.70\% on Librispeech test-clean and test-other set, respectively.
\end{abstract}
% VZ: I would skip e.g. and brackets in the abstract
% VZ: 4% to 11%
% VZ: comma before respectively
\begin{keywords}
end-to-end speech recognition, LVCSR, policy gradient, deep neural networks
\end{keywords}
\section{Introduction}
\label{sec:intro}
%Deep neural networks have become a popular choice for acoustic modeling in speech recognition with promising results \cite{hinton2012deep,saon2015ibm,xiong2017microsoft}. 
%RS: Never start with x is popular in your introduction!

Deep neural networks are the basis for some of the most accurate speech recognition systems in research and production \cite{hinton2012deep,saon2015ibm,xiong2017microsoft}. 
Neural network based acoustic models are commonly used as a sub-component in a Gaussian mixture model (GMM) and hidden Markov model (HMM) based hybrid system.
%NK: Is it correct to say "*is* commonly" given that more recent models use CNN-LSTMs or equivalent? For example, the work by Tara Sainath.
Alignment is necessary to train the acoustic model, and a two-stage (\ie alignment and frame prediction) training process is required for a typical hybrid system. A drawback of such setting is that there is a disconnect between the acoustic model training and the final objective, which makes the system level optimization difficult. 
%NK: which is......
% VZ: which makes optimization on the system level difficult.

The end-to-end neural network based speech models bypass this two-stage training process by directly maximizing the likelihood of the data. More recently, the end-to-end models have also shown promising results on various datasets \cite{graves2014towards,miao2015eesen,amodei2016deep,bahdanau2016end}. While the end-to-end models are commonly trained with maximum likelihood, the final performance metric for a speech recognition system is typically word error rate (WER) or character error rate (CER). This results a mismatch between the objective that is optimized and the evaluation metric. In an ideal setting the model should be trained to optimize the final metric. However, since the metrics are commonly discrete and non-differentiable, it is very difficult to optimize them in practice. %NK: optimize "them" in practice

Lately, reinforcement learning (RL) has shown to be effective on improving performance for problems that have non-differentiable metric through policy gradient. Promising results are obtained in machine translation \cite{ranzato2015sequence, bahdanau2016actor}, image captioning \cite{ranzato2015sequence, rennie2016self}, summarization \cite{ranzato2015sequence,paulus2017deep}, \etc.  In particular, REINFORCE algorithm \cite{williams1992simple} enables one to estimate the gradient of the expected reward by sampling from the model. It has also been applied for online speech recognition \cite{luo2017learning}. Graves and Jaitly \cite{graves2014towards} propose expected transcription loss that can be used to optimize on WER. However, it is more computationally expensive. For example, for a sequence of length $T$ with vocabulary size $K$, at least $T$ samples and $K$ metric calculations are required for estimating the loss.
%NK: $K$ metric calculations are required?
% VZ: I would remove the "e.g." here and write ". For example, "

We show that jointly training end-to-end models with self critical sequence training (SCST) \cite{rennie2016self} and maximum likelihood improves performance significantly. 
%NK: Two things. (a) Shouldn't it be "jointly training" given the usage of the word. (b) Split this sentence. It doesn't read well as one. 
SCST is also efficient during training, as only one sampling process and two metric calculations are necessary. Our model achieves 5.53\% WER on Wall Street Journal dataset, and 5.42\% and 14.70\% WER on Librispeech test-clean and test-other sets.
% VZ: Using SCST, our model achieves...

\begin{figure}[tbhp!]
\centering
\includegraphics[width=0.7\columnwidth,trim=2.1in 1.0in 4.8in 0.4in,clip]{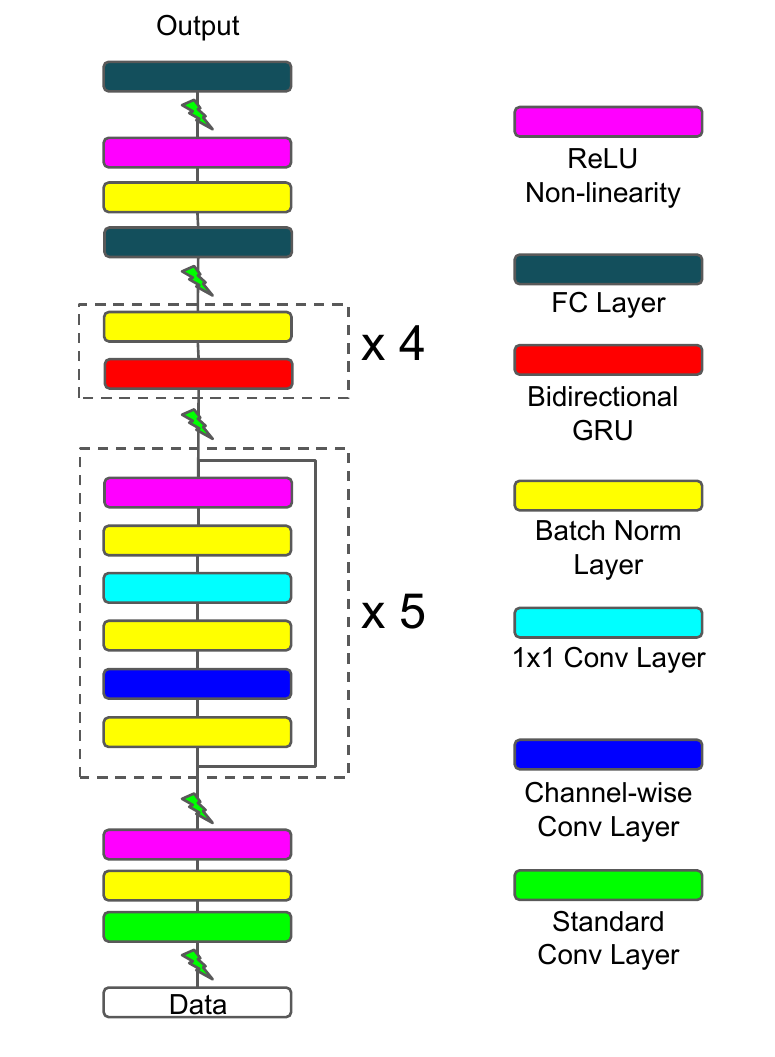}
\caption{Model architecture of our end-to-end speech model. Different colored blocks represent different layers as shown on the right, the lightning symbol indicates dropout happens between the two layers.}
\label{fig:model_structure}
\end{figure}

\section{Model Structure}
The end-to-end model structure used in this work is very similar to that of Deep Speech 2 (DS2) \cite{amodei2016deep}. It is mainly composed of 1) a stack of convolution layers in the front-end for feature extraction,  and 2) a stack of recurrent layers for sequence modeling. The structure of recurrent layers is the same as in DS2, and we illustrate the modifications in convolution layers in this section. 
% VZ: in the following what?

We choose to use time and frequency convolution (\ie 2-D convolution) as the front-end of our model, since it is able to model both the temporal transitions and spectral variations in speech utterances. We use depth-wise separable convolution \cite{sifre2013rotation, chollet2016xception} for all the convolution layers, due to its computational efficiency and performance advantage \cite{chollet2016xception}. The depth-wise separable convolution is implemented by first convolving over the input channel-wise, and then convolve with $1\times1$ filters with the desired number of output channels. Stride size only influences the channel-wise convolution; the following $1\times 1$ convolutions always have stride size of one. More precisely, let $\mathbf{x} \in \mathbb{R}^{F\times T\times D}$, $\mathbf{c} \in \mathbb{R}^{W \times H \times D}$ and $\mathbf{w} \in \mathbb{R}^{D \times N}$ denote an input sample, the channel-wise convolution and the $1\times 1$ convolution weights respectively. The depth-wise separable convolution with $D$ input channels and $N$ output channels performs the following operations:
\begin{align}
s(i,j,d) &= \sum_{f=0}^{F-1}\sum_{t=0}^{T-1} x(f,t,d) c(i-f,j-t,d) \\
o(i,j,n) &= \sum_{k=0}^{D-1} s(i,j,k) w(k,n) 
\end{align}
where $d \in\{1,\ldots,D\}$ and $n \in \{1,2,\ldots,N\}$, $\mathbf{s}$ is the channel-wise convolution result, and $\mathbf{o}$ is the result from depth-wise separable convolution.
In addition, we add a residual connection \cite{he2016deep} between the input and the layer output for the depth-wise separable convolution to facilitate training. 
% VZ: not sure what the "front end of the model" means
% VZ: not sure why you mention "stride size only influence the...", also it should be "influences"
% VZ: In more detail is awkward, "In particular" or just delete it.

Our model is composed of six convolution layers -- one standard convolution layer that has larger filter size, followed by five residual convolution blocks \cite{he2016deep}.
%\footnote{When the input only has one channel, depth-wise separable convolution is equivalent to standard convolution.}
% VZ: not sure why the footnote is necessary
%NK: Cite He et. al. here?
The convolution features are then fed to four bidirectional gated recurrent units (GRU) \cite{cho2014properties} layers, and finally two fully connected layers that make the final per-character prediction. %Batch normalization \cite{bn} is applied to all layers' pre-activations to facilitate training. 
%NK: pre-activation output?
%Dropout \cite{dropout} is applied to inputs of each layer, and for layers that take sequential input (\ie the convolution and recurrent layers) we use the dropout variant proposed by Gal and Ghahramani \cite{gal2016theoretically}. 
The full end-to-end model structure is illustrated in Fig. \ref{fig:model_structure}.
% VZ: these things read like they should be in the experiment details section. They are so detailed they detract from your contributions.

\section{Model Objective}
\subsection{Maximum Likelihood Training}
Connectionist temporal classification (CTC) \cite{graves2006ctc}
%NK: Why is the citation in the brackets? Seems weird to me. 
is a popular method for doing maximum likelihood training on sequence labeling tasks, where the alignment information is not provided in the label. The alignment is not required since CTC marginalizes over all possible alignments, and maximizes the likelihood $P(\mathbf{y}|\mathbf{x})$. It achieves this by augmenting the original label set $\mathcal{L}$ to set $\Omega = \mathcal{L} \cup \{\text{blank}\}$ with an additional blank symbol. A mapping $\mathcal{M}$ is then defined to map a length $T$ sequence of label $\Omega^T$ to $\mathcal{L}^{\le T}$ by removing all blanks and repeated symbols along the path. The likelihood can then be recovered by 
\begin{align*}
P(\mathbf{y}' | \mathbf{x}) &= \prod_t P( y'_t | \mathbf{x}), y'_t \in \Omega^T \\
P(\mathbf{y}|\mathbf{x}) &= \sum_{\mathbf{y}' \in \mathcal{M}^{-1}(\mathbf{y})} P(\mathbf{y}' | \mathbf{x})
\end{align*}
where $\mathbf{x}$, $\mathbf{y}$ and $\mathbf{y}'$ denote an input example of length $T$, the corresponding label of length $\le T$ and one of the augmented label with length $T$.
% VZ: I don't think \Omega has been introduced??
% VZ: We define the label set \Omega as the the union between the original label set \Omega^\prime \union blank, where blank [achieves the whatever goal]
% VZ: I don't know what it means to have \le in the superscript

\subsection{Policy Learning}
The log likelihood reflects the log probability of getting the whole transcription completely correct. What it ignores are the probabilities of the incorrect transcriptions. In other words, all incorrect transcriptions are equally bad, which is clearly not the case. Furthermore, the performance metrics typically aim to reflect the plausibility of incorrect predictions. For example, WER penalizes less for transcription that has less edit distance to the ground truth label.
%However, when the performance of a model is measured we care about both probabilities (\eg WER measures the number of insertions, deletions and substitutions when there are transcription errors). 
This results in a disparity between the optimization objective of the model and the (commonly discrete) evaluation criteria. This mismatch is mainly attributed to the inability to directly optimize the criteria.

One way to remedy this mismatch is to view the above problem in the policy learning framework. In this framework, we can view our model as an \emph{agent} and the training samples as the \emph{environment}. The parameters of the model $\theta$ defines a policy $P_\theta(\mathbf{y}|\mathbf{x})$, the model interacts with the environment by following this policy. The agent then performs an \emph{action} based on its current \emph{state}, in which case the action is the generated transcription and the state is the model hidden representation of the data. It then observes a \emph{reward} that is defined from the evaluation metric
%NK: There are too many 'in which case's here. 
calculated on the current sample (\eg $1 -$WER for the current transcription). The goal of learning is to obtain a policy that minimizes the negative expected reward:
\begin{equation}
\label{eq:policy}
L_p(\theta) = - \mathbb{E}_{\mathbf{y}^s\sim P_\theta(\mathbf{y}|\mathbf{x})}[r(\mathbf{y}^s)]
\end{equation}
where $r(\cdot)$ denotes the reward function. Gradient of eq. \ref{eq:policy} can be obtained through REINFORCE \cite{williams1992simple} as
\begin{align}
\label{eq:policy_grad}
\nabla_\theta L_p(\theta) &= - \mathbb{E}_{\mathbf{y}^s\sim P_\theta(\mathbf{y}|\mathbf{x})}[r(\mathbf{y}^s)\nabla_\theta \log P_\theta(\mathbf{y}^s|\mathbf{x})] \\
\label{eq:policy_grad_approx}
& \approx -r(\mathbf{y}^s)\nabla_\theta \log P_\theta(\mathbf{y}^s|\mathbf{x})
\end{align}
Eq. \ref{eq:policy_grad_approx} shows the Monte Carlo approximation of the gradient with a single example, which is a common practice when training model with stochastic gradient descent.
% VZ: the variables are not explained here. For example what is P\Theta(y|x), $y^s$ etc...

The policy gradient obtained from eq. \ref{eq:policy_grad_approx} is often of high variance, and the training can get unstable. To reduce the variance, Rennie \etal \cite{rennie2016self} proposed self-critical sequence training (SCST). In SCST, the policy gradient is computed with a \emph{baseline}, which is the greedy output from the model. Formally, the policy gradient is calculated using
\begin{align}
\nabla_\theta L_p(\theta) &= - \mathbb{E}_{\mathbf{y}^s\sim P_\theta(\mathbf{y}|\mathbf{x})}[\left(r(\mathbf{y}^s) - r(\hat{\mathbf{y}})\right) \nabla_\theta \log P_\theta(\mathbf{y}^s|\mathbf{x})] \\
\label{eq:scst_approx}
& \approx -\left(r(\mathbf{y}^s) - r(\hat{\mathbf{y}})\right) \nabla_\theta \log P_\theta(\mathbf{y}^s|\mathbf{x})
\end{align}
where $\hat{\mathbf{y}}$ is the greedy decoding output from the model for the input sample $\mathbf{x}$.
% VZ: I would avoid curly brace$. Use \left( and \right) to have auto-sizing brackets
% VZ: I think it'd be nice if you show the extra step of taking the reward out of the expectation, but that is a minor detail

\subsection{Multi-objective Policy Learning}
A potential problem with policy gradient methods (including SCST) is that the learning can be slow and unstable at the beginning of training. This is because it is unlikely for the model to have reasonable output at that stage, which leads to implausible samples with low rewards. Learning will be slow in case of small learning rate, and unstable otherwise. One way to remedy this problem is to incorporate maximum likelihood objective along with policy gradient, since in maximum likelihood the probability is evaluated on the ground truth targets, and hence will get large gradients when the model output is incorrect. This leads to the following objective for training our end-to-end speech model:
\begin{align}
\label{eq:model_obj}
L(\theta) = & - \log P_\theta(\mathbf{y}|\mathbf{x}) + \lambda L_{scst}(\theta) \qquad \text{where}\\
\nonumber
L_{scst}(\theta) = & - \{g(\mathbf{y}^s, \mathbf{y}) - g(\hat{\mathbf{y}}, \mathbf{y})\}\log P_\theta(\mathbf{y}^s|\mathbf{x})
\end{align}
where $g(\cdot,\cdot)$ is the reward function and $\lambda \in (0, +\infty)$ is the coefficient that controls the contribution from SCST. In our case we choose $g(\cdot, \mathbf{y}) = 1 - \max(1, \text{WER}(\cdot, \mathbf{y}))$. Training with eq. \ref{eq:model_obj} is also efficient, since both sampling and greedy decoding is cheap. The only place that might be computationally more demanding is the reward calculation, however, we only need to compute it twice per batch of examples, which adds only a minimal overhead.
% VZ: Why is it unlikely? (because the policy learning has just started and the current policy sucks)
% VZ: because you put the equations on two lines, it seems like they are equal to each other, which is not the case because they are two objectives.

\section{Experiments}
\label{sec:exp}
We evaluate the proposed objective by performing experiments on the Wall Street Journal (WSJ) and LibriSpeech \cite{panayotov2015librispeech} datasets. The input to the model is a spectrogram computed with a 20ms window and 10ms step size.
We first normalize each spectrogram to have zero mean and unit variance. In addition, we also normalize each feature to have zero mean and unit variance based on the training set statistics. No further preprocessing is done after these two steps of normalization. 
%NK: I forget (& this might be a stupid question) but does the order matter here?

We denote the size of the convolution layer by the tuple $(\text{C, F, T, SF, ST})$, where C, F, T, SF, and ST denote number of channels, filter size in frequency dimension, filter size in time dimension, stride in frequency dimension and stride in time dimension respectively. We have one convolutional layer with size (32,41,11,2,2), and five residual convolution blocks of size (32,7,3,1,1), (32,5,3,1,1), (32,3,3,1,1), (64,3,3,2,1), (64,3,3,1,1) respectively. Following
the convolutional layers we have 4 layers of bidirectional GRU RNNs with $1024$ hidden units per direction per layer. Finally, we have one fully connected hidden layer
of size $1024$ followed by the output layer. Batch normalization \cite{bn} is applied to all layers' pre-activations to facilitate training. 
Dropout \cite{dropout} is applied to inputs of each layer, and for layers that take sequential input (\ie the convolution and recurrent layers) we use the dropout variant proposed by Gal and Ghahramani \cite{gal2016theoretically}. The convolutional and fully connected layers are initialized uniformly following He \etal \cite{he2015delving}. The recurrent layer weights are initialized with a uniform distribution $\mathcal{U}(-1/32, 1/32)$. The model is trained in an end-to-end fashion to minimize the mixed objective as illustrated in eq. \ref{eq:model_obj}. We use mini-batch stochastic gradient descent with batch size 64, learning rate 0.1, and with Nesterov momentum 0.95. The learning rate is reduced by half whenever the validation loss has plateaued. 
We set $\lambda = 0.1$ at the beginning of training, and increase it to $1$ after the model has converged (\ie the validation loss stops improving). 
The gradient is clipped \cite{pascanu2013difficulty} to have a maximum $\ell_2$ norm of $1$.
%NK: Which norm? Infinity or L2?
For regularization, we use $\ell_2$ weight decay of $10^{-5}$ for all 
%NK: I usually prefer $\ell_2$ but it is up to you. 
parameters. Additionally, we apply dropout for inputs of each layer (see Fig. \ref{fig:model_structure}). The dropout probabilities are set as $0.1$ for data, $0.2$ for all convolution layers, and $0.3$ for all recurrent and fully connected layers. Furthermore, we also augment the audio training data through random perturbations of tempo, pitch, volume, temporal alignment, along with adding random noise. 
%NK: Is this a contribution of your other paper? Maybe cite it here as an unplublished manuscript? Or you can always do that in the camera ready.

\subsection{Effect of Policy Learning}
To study the effectiveness of our multi-objective policy learning, we perform experiments on both datasets with various settings. The first set of experiments was carried out on the WSJ corpus. We use the standard \emph{si284} set for training, \emph{dev93} for validation and \emph{eval92} for test evaluation. We use the provided language model and report the result in the 20K closed vocabulary setting with beam search. The beam width is set to 100. Results are shown in table \ref{tbl:wsj_comp}. Both policy gradient methods improve results over baseline. In particular, the use of SCST results in 13.8\% relative performance improvement on the \emph{eval92} set over the baseline.

\begin{table}[tbp!]
\centering
\begin{tabular}{lcccc}
\toprule
\multirow{2}{*}{Method} & \multicolumn{2}{c}{dev93} & \multicolumn{2}{c}{eval92} \\
						& CER & WER & CER & WER \\
\midrule
Baseline 								 & 4.07\% & 9.93\% & 2.59\% & 6.42\%\\
Policy (eq. \ref{eq:policy_grad_approx}) & 3.71\% & 9.46\% & 2.31\% & 5.85\%\\
Policy (eq. \ref{eq:scst_approx})  		 & 3.52\% & 9.21\% & 2.10\% & 5.53\%\\
\bottomrule
\end{tabular}
\caption{Performance from WSJ dataset. Baseline denotes model trained without CTC only; policy indicates model trained using the multi-objective policy learning. Equation in parenthesis indicates the way used to obtain policy gradient.}
\label{tbl:wsj_comp}
\end{table}
% would be nice to have line dividing CER, WER and dev93 etc.

On LibriSpeech dataset, the model is trained using all 960 hours of training data. Both dev-clean and dev-other are used for validation and results are reported in table \ref{tbl:libri_comp}. The provided 4-gram language model is used for final beam search decoding. The beam width is also set to 100 for decoding. Overall, a relative $\approx 4\%$ performance improvement over the baseline is observed.
%VZ: don't need the approx - just round it

\begin{table}[tbp!]
\centering
\begin{tabular}{lccc}
\toprule
\multicolumn{2}{c}{Dataset} & Baseline & Policy \\
\midrule
\multirow{2}{*}{dev-clean} 	& CER & 1.76\% 	& 1.69\%\\
							& WER & 5.33\%	& 5.10\%\\
\midrule
\multirow{2}{*}{test-clean} & CER & 1.87\% 	& 1.75\%\\
							& WER & 5.67\%	& 5.42\%\\
\midrule
\multirow{2}{*}{dev-other} 	& CER & 6.60\% 	& 6.26\%\\
							& WER & 14.88\% & 14.26\%\\
\midrule
\multirow{2}{*}{test-other} & CER & 6.58\% 	& 6.25\%\\
							& WER & 15.18\%	& 14.70\%\\
\bottomrule
\end{tabular}
\caption{Performance from LibriSpeech dataset. Policy denotes model trained with multi-objective shown in eq. \ref{eq:model_obj}.}
\label{tbl:libri_comp}
\end{table}

\begin{table}[tbp!]
\centering
\begin{tabular}{lcc}
\toprule
Method  & WER \\
\midrule
Hannun \etal \cite{hannun2014first} & 14.10 \%\\
Bahdanau \etal \cite{bahdanau2016end} & 9.30\%\\
Graves and Jaitly \cite{graves2014towards} & 8.20\% \\
Wu \etal \cite{wu2016multiplicative} & 8.20\% \\
Miao \etal \cite{miao2015eesen} & 7.34\% \\
Chorowski and Jaitly \cite{chorowski2016towards} & 6.70\%\\
Human \cite{amodei2016deep} & 5.03\% \\
Amodei \etal \cite{amodei2016deep}* & 3.60\% \\
Ours & 5.53\%\\
Ours (LibriSpeech) & 4.67\%\\
\bottomrule
\end{tabular}
\caption{Comparative results with other end-to-end methods on WSJ \emph{eval92} dataset. LibriSpeech denotes model trained using LibriSpeech dataset \emph{only}, and test on WSJ. Amodei \etal used more training data.}
\label{tbl:wsj}
\end{table}

\begin{table}[tbp!]
\centering
\begin{tabular}{lcc}
\toprule
Method & test-clean & test-other \\
\midrule
Collobert \etal \cite{collobert2016wav2letter} & 7.20\% & - \\
Amodei \etal \cite{amodei2016deep}* & 5.33\% & 13.25\% \\
ours & 5.42\% & 14.70\% \\
\bottomrule
\end{tabular}
\caption{Word error rate comparison with other end-to-end methods on LibriSpeech dataset. Amodei \etal used more training data.}
\label{tbl:libri}
\end{table}

\subsection{Comparison with Other Methods}
We also compare our performance with other end-to-end models. Comparative results from WSJ and LibriSpeech dataset are illustrated in tables \ref{tbl:wsj} and \ref{tbl:libri} respectively.
%NK: tables .... respectively
Our model achieved competitive performance with other methods on both datasets. In particular, with the help of policy learning we achieved similar results as Amodei \etal \cite{amodei2016deep} on LibriSpeech without using additional data. To see if the model generalizes, we also tested our LibriSpeech model on the WSJ dataset. The result is significantly better than the model trained on WSJ data (see table \ref{tbl:wsj}), which suggests that the end-to-end models benefit more when more data is available. 

\section{Conclusion}
In this work, we try to close the gap between the maximum likelihood training objective and the final performance metric for end-to-end speech models. We show this gap can be reduced by using the policy gradient method along with the negative log-likelihood. In particular, we apply a multi-objective training with SCST to reduce the expected negative reward that is defined by using the final metric. The joint training is computationally efficient. We show that the joint training is effective even with single sample approximation, which improves the relative performance on WSJ and LibriSpeech by 13\% and 4\% over the baseline. 

\bibliographystyle{IEEEbib}
\bibliography{refs}

\end{document}